\documentclass[11pt]{article}

\usepackage[final]{acl}

\usepackage{times}
\usepackage{amsmath}
\usepackage{amssymb}
\usepackage{latexsym}
\usepackage{makecell}
\usepackage{booktabs}
\usepackage{multirow}
\usepackage{graphicx}
\usepackage{xcolor}
\usepackage{booktabs}
\usepackage{colortbl}
\usepackage[table]{xcolor}
\usepackage[utf8]{inputenc}    
\usepackage{tabularx}
\usepackage[table]{xcolor}
\usepackage{booktabs}
\usepackage{multirow} 
\usepackage{makecell} 
\usepackage[T1]{fontenc}

\usepackage[utf8]{inputenc}

\usepackage{microtype}

\usepackage{inconsolata}

\usepackage{graphicx}
\definecolor{lightblue}{RGB}{220,230,255}
\title{State-Conditioned Visual Evidence Retrieval for Fine-Grained Perception in Document Vision-Language Models}

\author{
\textbf{Mingxu Chai}\textsuperscript{1,2,3}\thanks{Equal contribution.}, 
\textbf{Chenyu Liu}\textsuperscript{1}\footnotemark[1], 
\textbf{Ziyu Shen}\textsuperscript{1}, 
\textbf{Jiazheng Zhang}\textsuperscript{1}, \\
\textbf{Kaidi Zhang}\textsuperscript{4}, 
\textbf{Ruoyu Chen}\textsuperscript{4},  
\textbf{Jun Long}\textsuperscript{4},  
\textbf{Jihua Kang}\textsuperscript{4},  
\textbf{Tao Gui}\textsuperscript{1,2}, 
\textbf{Qi Zhang}\textsuperscript{1,3,5}\thanks{Corresponding author.} \\
\textsuperscript{1}College of Computer Science and Artificial Intelligence, Fudan University \\
\textsuperscript{2}Shanghai Innovation Institute
\textsuperscript{3}Shanghai Artificial Intelligence Laboratory \\
\textsuperscript{4}ByteDance, LarkAI
\textsuperscript{5}Shanghai Key Lab of Intelligent Information Processing \\
\texttt{qz@fudan.edu.cn}
}

\begin{document}
\maketitle
\begingroup
\renewcommand\thefootnote{}
\footnotetext{Code: \url{https://github.com/SII-sc22mc/SCVER}.}
\addtocounter{footnote}{-1}
\endgroup

\begin{abstract}
Compared with typical vision–language tasks, document parsing places stronger demands on fine-grained visual perception. 
Existing vision–language model (VLM)–based parsing approaches rely on globally compressed visual tokens, where fine-grained details are entangled within a single representation and repeatedly accessed during decoding.
However, we observe that the visual evidence for each prediction is typically localized and conditioned on the current decoding state, whereas such representations must be accessed in full at every decoding step, resulting in inefficient computation.
To address this mismatch, we formulate perception as state-conditioned visual evidence retrieval (SCVER) during autoregressive decoding. The model operates on a compact global representation for coarse structure and retrieves a small set of relevant high-resolution regions conditioned on the current token state. This coarse-to-fine design enables on-demand access to fine-grained visual cues, relieving globally shared representations from encoding all fine-grained details.
We further find that learning such state-conditioned retrieval in VLMs is challenging and unstable. To stabilize this process, we introduce a Spatially-Guided Learning Objective (SGLO) to guide the retrieval process.
Experiments on document parsing benchmarks show that SCVER improves robustness under reduced input resolution and achieves a better accuracy–efficiency trade-off, demonstrating the effectiveness of on-demand visual evidence retrieval for fine-grained perception.
\end{abstract}

\begin{figure}[t]
  \centering
  \includegraphics[width=1.0\linewidth, trim= 10 273 550 0, clip]{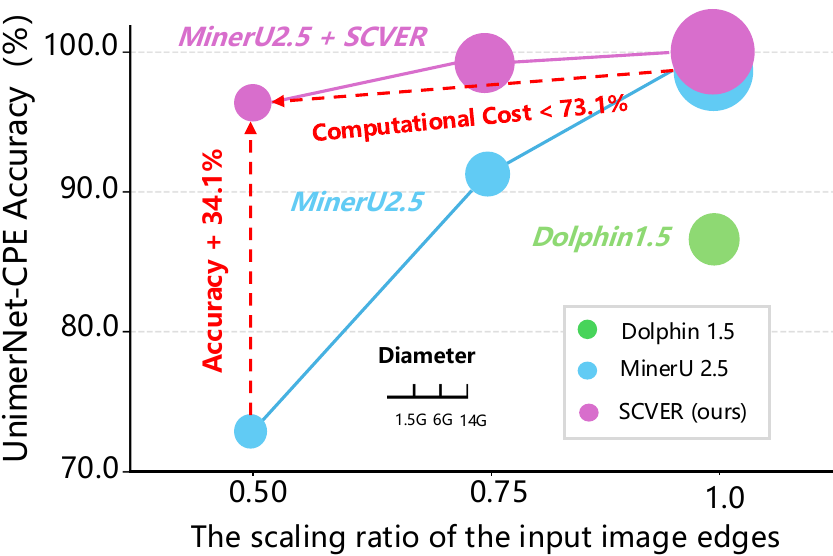}
  \caption{\textbf{Accuracy vs. Input Size on UniMER-CPE}~\cite{wang2024unimernetuniversalnetworkrealworld}. Circle size indicates the average GFLOPs per generated token. SCVER reduces the reliance on high-resolution inputs, achieving a more favorable accuracy–efficiency trade-off compared to the dynamic-resolution MinerU2.5~\cite{niu2025mineru25decoupledvisionlanguagemodel} and the fixed-resolution Dolphin~\cite{feng2025dolphindocumentimageparsing}.}
\label{fig:1}
\end{figure}
\section{Introduction}

\begin{figure*}
  \centering
  \includegraphics[width=1.0\linewidth, trim=0 285 0 0, clip]{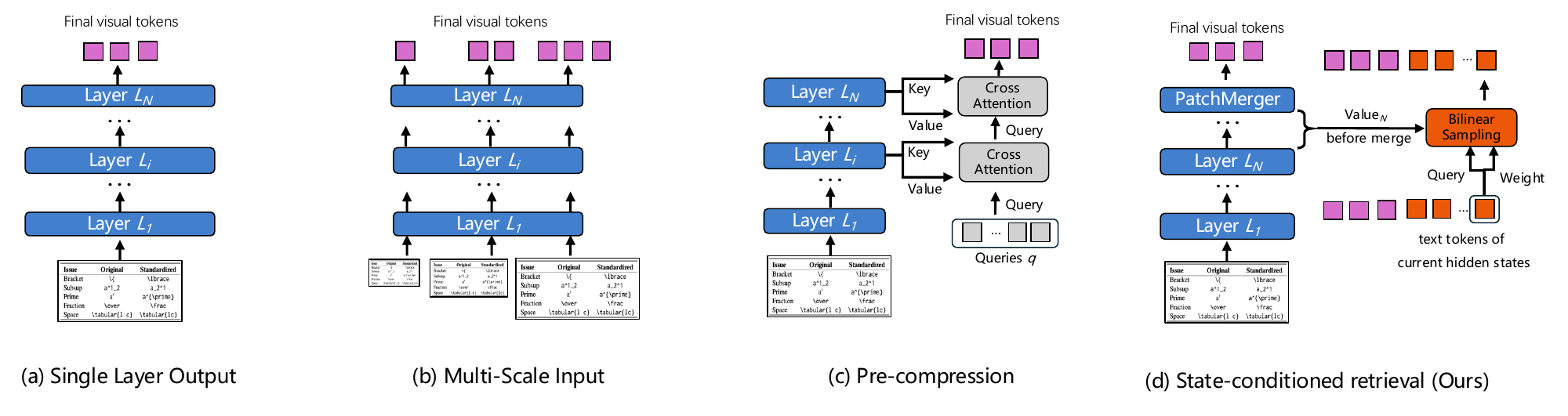}
  \caption{(a) Most VLMs only use final layer output. (b) LLaMA-Adapter~\cite{zhang2024llamaadapterefficientfinetuninglanguage} injects multi-scale visual features into the decoder. (c) VIVL\cite{Cao_2024_CVPR} uses a small set of fixed queries to gather multi-level features, but these sparse visual samples stay constant throughout decoding, offering limited information. (d)Taking MinerU2.5\cite{niu2025mineru25decoupledvisionlanguagemodel} as an example, SCVER dynamically supplies pre-downsampling visual information required at each decoding step.}
  \label{fig:intro2}
\end{figure*}

Document parsing aims to convert document images into machine-readable formats \cite{cui2025paddleocrvlboostingmultilingualdocument}. Unlike typical vision–language (VL) tasks that primarily focus on cross-modal alignment and semantic reasoning, document parsing relies more on precise visual perception and fine-grained feature modeling \cite{wei2024generalocrtheoryocr20}.

Despite recent advances in VLM-based document parsing, most methods encode the entire image into a fixed set of visual tokens and rely on them throughout decoding \cite{niu2025mineru25decoupledvisionlanguagemodel}. This forces a single global representation to capture both coarse semantics and fine-grained details, while requiring full interactions with all visual tokens at every decoding step.
However, we observe that the visual evidence required for prediction is typically dynamic and spatially localized, with each step depending on only a small subset of regions. This creates a fundamental mismatch between localized, step-dependent evidence and global, fully accessed representations, leading to redundant computation.
The problem further worsens as input resolution increases, since the number of visual tokens grows, making repeated full interactions increasingly expensive under quadratic-cost attention \cite{zhang2025documentparsingunveiledtechniques}.

Based on this observation, we reformulate visual perception as state-conditioned visual evidence retrieval (SCVER) during decoding, enabling on-demand access to fine-grained information. Instead of requiring a single global representation to encode all visual details upfront, the model first operates on compressed visual tokens to capture coarse structure and establish global text–vision alignment, and then dynamically retrieves a small set of relevant high-resolution regions conditioned on the current decoding state (Fig.~\ref{fig:intro2}(d)).
As a result, global visual tokens are no longer required to preserve all fine-grained details, in contrast to conventional VLMs and multi-scale approaches that rely on pre-aggregated representations (Fig.~\ref{fig:intro2}(a–c)). This design reduces the dependence on high-resolution inputs while maintaining efficient decoding.

However, we find that naively applying SCVER leads to unstable optimization and limited gains. The difficulty stems from the absence of explicit spatial supervision in state-conditioned retrieval. While standard attention grounds relevance through query–key similarity, SCVER must predict retrieval locations directly, introducing latent spatial variables that are only indirectly supervised by token-level losses. This makes the spatial selection process underconstrained and prone to degenerate behaviors.
To address this issue, we introduce a Spatially-Guided Learning Objective (SGLO), which provides explicit spatial guidance by leveraging the visual-token attention from an independent frozen-base
forward as a weak spatial prior over relevant regions, thereby stabilizing the retrieval process.


Our experiments show that SCVER consistently improves both efficiency and robustness, especially under reduced input resolution. When integrated with MinerU2.5~\cite{niu2025mineru25decoupledvisionlanguagemodel}, it reduces computation by over 70\% with negligible accuracy loss, leading to an improved accuracy–efficiency trade-off (Fig.~\ref{fig:1}). Moreover, SCVER significantly mitigates performance degradation under constrained resolutions, for example improving PubTabNet~\cite{pubtab} accuracy from 86.9 to 91.2 on MonkeyOCR~\cite{li2025monkeyocrdocumentparsingstructurerecognitionrelation}.
The main contributions are summarized as follows:

\begin{itemize}
    \item We identify a mismatch between localized, state-dependent visual evidence and global representations in document VLMs.

    \item We propose SCVER, a decoding-time mechanism that enables on-demand, token-specific access to fine-grained visual evidence.

    \item We introduce the SGLO to stabilize retrieval learning without additional annotations.
\end{itemize}

\section{Related Works}
\label{sec:Related Works}

\subsection{Document Parsing Models}
Document parsing involves layout detection and various recognition tasks. Early approaches integrated multiple specialized models\cite{zhao2024doclayoutyoloenhancingdocumentlayout, wang2024unimernetuniversalnetworkrealworld,da2023visiongridtransformerdocument} into modular pipeline systems\cite{wang2024mineruopensourcesolutionprecise}.
To reduce system complexity, research gradually shifted toward end-to-end approaches that directly map document images to structured text\cite{blecher2023nougatneuralopticalunderstanding, wei2024generalocrtheoryocr20}.
In recent years, the document parsing paradigm has evolved into a lightweight pipeline that first detects and then recognizes in parallel \cite{feng2025dolphindocumentimageparsing,cui2025paddleocrvlboostingmultilingualdocument}.
However, these works mainly focus on the overall parsing paradigm, with limited attention to fine-grained visual information utilization during decoding.

\subsection{Visual Information Utilization in VLMs}
Dynamic-resolution encoders, such as NaViT, process images with flexible patch resolutions to handle aspect ratio variations \cite{dehghani2023patchnpacknavit}, but do not explicitly enhance fine-grained perception.
Multi-scale approaches aggregate features before decoding \cite{park2024hierarchicalvisualfeatureaggregation}, improving representation capacity at the cost of efficiency. 
Recent work makes visual access adaptive to intermediate states, but remains within the standard attention framework, where visual evidence is implicitly aggregated rather than explicitly retrieved in a state-dependent manner.
Some methods also incorporate external OCR modules \cite{nacson2024docvlmmakevlmefficient}. Overall, these approaches do not explicitly enable state-dependent, selective access to fine-grained visual evidence during decoding.


\subsection{Sparse Attention Mechanisms}
In vision tasks, deformable attention samples a small set of spatial locations for each query to enable efficient multi-scale feature aggregation \cite{zhu2021deformabledetrdeformabletransformers}.
To reduce the quadratic cost of dense attention, prior work explores sparse attention by restricting interactions to subsets of tokens, including predefined patterns and similarity-based selection \cite{kitaev2020reformerefficienttransformer, roy2020efficientcontentbasedsparseattention, beltagy2020longformerlongdocumenttransformer, zaheer2021bigbirdtransformerslonger}. Recent work further improves efficiency through token pruning, such as SparseVLM \cite{zhang2025sparsevlmvisualtokensparsification}, which reduces redundant visual tokens.
In contrast, our method enables state-dependent visual access by explicitly retrieving relevant visual evidence conditioned on the current decoding state.
\section{Method}

\subsection{State-Conditioned Visual Evidence Retrieval}

Given an input image $I \in \mathbb{R}^{H \times W \times 3}$, the visual encoder produces high-resolution features $\tilde{F} \in \mathbb{R}^{\tilde{n}_v \times \tilde{d}}$. To reduce cost, a Patch Merger compresses them into $F \in \mathbb{R}^{n_v \times d}$ with $n_v < \tilde{n}_v$. The compressed visual tokens are concatenated with textual embeddings $T \in \mathbb{R}^{n_t \times d}$ to form $X_0 = [F; T]$, which is processed by a Transformer decoder.

\begin{figure}[t]
  \centering
  \includegraphics[width=1.0\linewidth, trim= 0 100 570 0, clip]{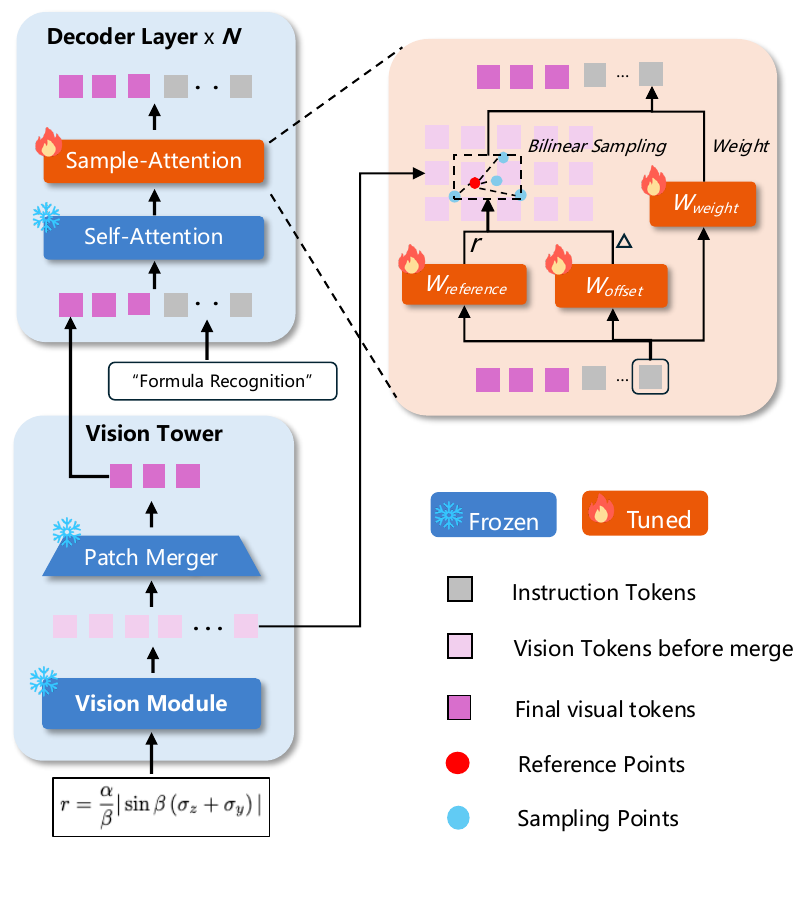}
  \caption{\textbf{Illustration of State-Conditioned Visual Evidence Retrieval.} SCVER efficiently retrieves fine-grained visual information from pre-downsampling features conditioned on the current decoding state.}
\label{fig:2}
\end{figure}

Most existing approaches rely on this compressed representation throughout decoding, requiring it to encode both global semantics and fine-grained details. This leads to a trade-off: increasing $n_v$ preserves more detail but incurs quadratic attention cost $\mathcal{O}((n_v + n_t)^2)$, while reducing $n_v$ improves efficiency at the expense of fine-grained information.

However, in document parsing, the visual evidence required for each prediction is typically localized and varies across decoding steps. This suggests that not all visual information needs to be pre-aggregated into a fixed representation.

Based on this observation, we formulate visual perception as a state-conditioned visual evidence retrieval problem. Instead of relying solely on $F$, the decoder dynamically retrieves relevant visual cues from $\tilde{F}$ conditioned on its current hidden state.

Formally, at each decoder block $i$, we augment the Transformer as:
\begin{equation}
\begin{aligned}
H_i &= X_{i-1} + \text{Attn}(X_{i-1}), \\
\tilde{H}_i &= \text{SCVER}(H_i, \tilde{F}), \\
X_i &= \tilde{H}_i + \text{FFN}(\tilde{H}_i),
\end{aligned}
\end{equation}

SCVER enables the model to access token-specific, high-resolution visual evidence on demand, rather than requiring all details to be encoded in the compressed tokens. As a result, the decoder can maintain a compact global representation while retrieving fine-grained information only when needed, improving both efficiency and information fidelity.

\subsection{Implementation of SCVER}
\label{scver_impl}

To instantiate SCVER, we design a lightweight retrieval operator that selectively extracts localized visual evidence from the high-resolution feature map $\tilde{F}$. The retrieval is conditioned on the token representation, allowing different decoding states to access different spatial regions.


Our implementation builds on the sparse sampling formulation of deformable attention~\cite{zhu2021deformabledetrdeformabletransformers}. In SCVER, the sampling parameters are conditioned on the current autoregressive token representation, allowing the retrieved visual evidence to dynamically vary with the decoding state (Fig.~\ref{fig:2} right).

Given a token $t \in \mathbb{R}^{d}$, we predict a reference point, offsets, and weights:
\begin{equation}
r = \sigma(t W_{ref}), \quad
\Delta = t W_{off}, \quad
w = t W_{weight},
\end{equation}
where $W_{ref} \in \mathbb{R}^{d \times 2}$ predicts a normalized reference point,
$W_{off} \in \mathbb{R}^{d \times (H \times P \times 2)}$ predicts offsets,
and $W_{weight} \in \mathbb{R}^{d \times (H \times P)}$ predicts sampling weights.
Here $H$ is the number of heads and $P$ the number of sampling points per head.

The offsets are reshaped to $\mathbb{R}^{H \times P \times 2}$ and normalized by $(W_f, H_f)$:
\begin{equation}
\ell_{h,p} = r + \frac{\Delta_{h,p}}{(W_f, H_f)}.
\end{equation}

For each head $h$, we bilinearly sample $\tilde{F}$ at $\ell_{h,p}$ and aggregate:
\begin{equation}
t'_h = \sum_{p=1}^{P}
w_{h,p}\, \mathrm{Bilinear}(\tilde{F}, \ell_{h,p}),
\end{equation}

The outputs are concatenated across heads as $t' = \mathrm{Concat}(t'_1, \dots, t'_H)$ and combined with the original token via a residual connection, enabling token-specific access to fine-grained visual evidence with minimal overhead while remaining compatible with KV caching.

\begin{table*}[ht]
\centering
\resizebox{\textwidth}{!}{%
\begin{tabular}{l c c c c c c c c c c c c c}
\toprule

\multirow{2}{*}{\textbf{Method}} & \multirow{2}{*}{\textbf{R}} & \multirow{2}{*}{\textbf{Param}} &
\multicolumn{4}{c}{Formula$^{\mathbf{CDM} \uparrow}$} &
\multicolumn{2}{c}{Table$^{\mathbf{TEDS} \uparrow}$} &
\multicolumn{2}{c}{Text$^{\mathbf{1-Edit} \uparrow}$}&
Page-Level \\
\cmidrule(lr){4-7}  \cmidrule(lr){8-9} \cmidrule(lr){10-11} \cmidrule(lr){12-12}
& & & SPE& SCE & CPE & HWE & PubTabNet & FinTabNet & DocLaynet & B-MOD & OmniDoc\\
\midrule
\multicolumn{10}{l}{\textit{General VLMs}} \\
\midrule
GPT-4o   & 1.0 & -        & 96.1 & 88.2 & 83.3 & 85.9 & 76.6 & 83.9 & 91.7& 0.318  &75.0 \\
Qwen3-VL-235B & 1.0 & 235B &98.4 & 96.2 & 97.4 & 94.2 & 85.4 & 84.9 & 92.3 & 0.321 &89.7\\
\midrule

\multicolumn{10}{l}{\textit{Specialized VLMs}} \\
\midrule
DeepSeek-OCR-2 &1.0&3B & 95.5 & 77.1 & 92.0& 81.6 & - & - & 92.1 & 81.2 &90.2         \\

GLM-OCR &1.0&0.9B & 98.4 & 97.7 & 96.7& 95.1 & 85.2 & 92.1 & 93.3 & 84.5 &95.2         \\

PaddleOCR-VL-1.5 &1.0&0.9B & 99.2 & 94.9 & 98.8& 92.3 & 84.6 & 94.7 & 93.6 & 85.1 &94.3         \\

\midrule
Dolphin-1.5&1.0&0.3B & 97.8 & 95.1 & 87.7& - & 85.9 & 87.4 & 91.4 & 82.7 &86.5         \\
\rowcolor{lightblue}
Dolphin-1.5+SCVER &1.0&0.3B  & 98.4 & 97.6 & 90.2& - & 91.9 & 92.4 & 92.6 & 94.0 &87.4         \\

\midrule
MonkeyOCR-Pro &1.0&3B & 97.6 & 94.9 & 91.4 & 92.2 & 87.4 & 86.4 & 92.0 & 83.3 &88.9 \\
MonkeyOCR-Pro &0.5&3B & 75.6 & 64.3 & 74.2 & 72.6 & 58.3 & 52.9 & 77.3 & 62.4 &71.3 \\
\rowcolor{lightblue}
MonkeyOCR-Pro+SCVER &0.5 &3B & 98.1 & 94.5 & 91.0 & 96.0 & 92.8 & 89.7 & 91.3 & 91.3 &89.1 \\
MonkeyOCR-Pro &0.7&3B & 94.5 & 92.1 & 88.6 & 85.4 & 79.6 & 78.1 & 86.3 & 75.9 &83.0 \\
\rowcolor{lightblue}
MonkeyOCR-Pro+SCVER  &0.7 &3B & 99.2 & \textbf{98.7} & 95.1 & \textbf{97.4} & \textbf{95.2} & 91.4 & 92.5 & \textbf{95.1} &90.7 \\

\midrule
MinerU2.5-Pro &1.0 &1.2B &  99.4 & 97.0 & 98.9& 95.3 & 90.1 & 95.1 & 92.9 &85.6 & 95.7 \\
MinerU2.5-Pro &0.5 &1.2B & 72.7 & 69.3 & 74.5& 65.7 & 66.5 & 73.6 & 74.5 &66.4 & 78.3 \\
\rowcolor{lightblue}
MinerU2.5-Pro+SCVER &0.5 &1.2B &  99.1 & 97.5 & 98.9& 96.1 & 90.4 & 95.7 & 92.3 &92.4 & 95.5 \\
MinerU2.5-Pro &0.7 &1.2B &  95.4 & 94.3 & 93.7& 90.4 & 82.0 & 89.1 & 86.8 &73.3 & 89.2 \\
\rowcolor{lightblue}
MinerU2.5-Pro+SCVER    &0.7 &1.2B  & \textbf{99.5} & 98.3 & \textbf{99.1}& 97.2 & 95.0 & \textbf{96.9} & \textbf{93.7} &94.7 & \textbf{95.9} \\
\bottomrule

\end{tabular}%
} 
\caption{Performance of different methods under varying image resolutions. 
R denotes the down-scaling ratio applied to each image dimension (width and height), corresponding to $R^2$ of the original image area. 
Values are approximate, as some models require padding or rounding to match implementation constraints. 
Bold numbers indicate the best results in each column.}
\label{tab:recognition}
\end{table*}

\begin{table}[ht]
\centering
\resizebox{\columnwidth}{!}{
\begin{tabular}{lccc}
\toprule
Model & Params & Encoder Type & Resolution \\
\midrule
Dolphin1.5\citeyearpar{feng2025dolphindocumentimageparsing} & 394M & Swin & Fixed \\
MinerU2.5\citeyearpar{niu2025mineru25decoupledvisionlanguagemodel} & 1.2B & ViT & Dynamic \\
MonkeyOCR-Pro\citeyearpar{li2025monkeyocrdocumentparsingstructurerecognitionrelation} & 3B & ViT & Dynamic \\
\bottomrule
\end{tabular}
}
\caption{Base models used in our experiments.}
\label{tab:baseline}
\end{table}

\subsection{Spatially-Guided Learning Objective}
\label{sec:SGLO}

We find that naively applying SCVER leads to unstable training and slow convergence. This difficulty arises from a fundamental limitation of state-conditioned visual retrieval: the retrieval locations are predicted directly from token representations, introducing latent spatial variables that determine where visual evidence is accessed. Unlike standard attention, where relevance is explicitly established through query--key similarity ($QK^\top$), these spatial variables are only weakly constrained by token-level supervision, making retrieval prone to diffuse or degenerate behaviors.

\begin{figure}[t]
  \centering
  \includegraphics[width=1.0\linewidth, trim=0 240 585 0, clip]{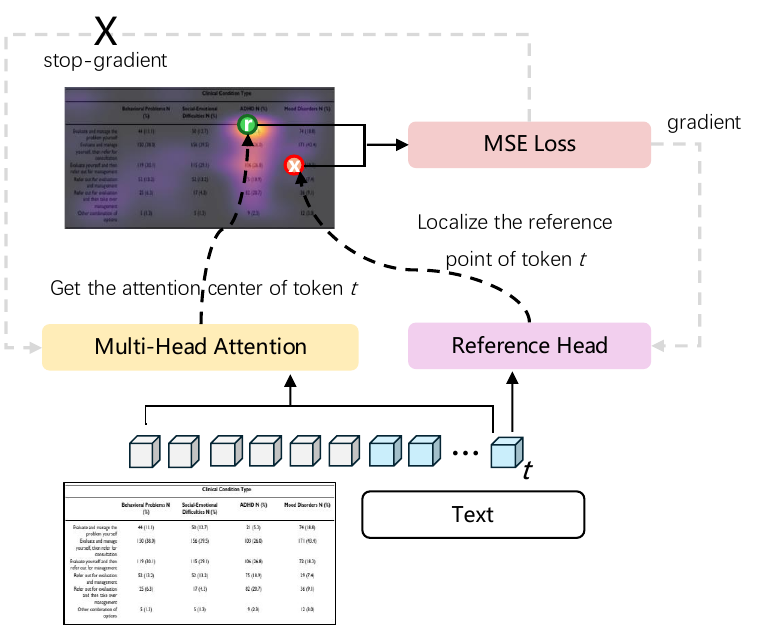}
  \caption{SGLO with two shared-weight hidden-state streams. 
The frozen base stream produces detached attention-centroid targets, 
while the SCVER stream predicts the corresponding reference points.}
  \label{fig:4}
\end{figure}

To address this, we introduce a Spatially-Guided Learning Objective (SGLO) that provides explicit spatial guidance for retrieval. During training, we maintain two hidden-state streams from the same input. A no-gradient base stream disables SCVER and preserves the pretrained hidden-state trajectory, whereas an independent SCVER stream enables the retrieval modules. Both streams share the same frozen backbone weights. We use the visual-token attention from the base stream as an online weak spatial prior, since it typically highlights regions informative for predicting each token.

At each SCVER insertion layer $i$, we use the causal self-attention
weights computed from the frozen-base hidden states
$X_{i-1}^{\mathrm{B}}$. (Fig.~\ref{fig:4}). For a textual token $t$, let $\alpha_t^{\mathrm{B}} \in \mathbb{R}^{H \times (n_v + n_t)}$ denote its multi-head attention weights. We extract the weights corresponding to visual tokens and average over heads:
\begin{equation}
\bar{\alpha}_t^{\mathrm{B}}
=
\frac{1}{H}
\sum_{h=1}^{H}
\alpha_t^{\mathrm{B},(h)}[1:n_v],
\end{equation}
which is reshaped into a spatial attention map $A_t^{\mathrm{B}} \in \mathbb{R}^{H_f \times W_f}$.

We compute its spatial centroid:
\begin{equation}
c_t^{\mathrm{B}}
=
\sum_{i,j}
A_t^{\mathrm{B}}(i,j)\,(x_{ij}, y_{ij}),
\end{equation}
where $(x_{ij}, y_{ij}) \in [0,1]^2$ are normalized coordinates.

In the SCVER stream, the corresponding hidden state $t^{\mathrm{S}}$ predicts the reference point $r^{\mathrm{S}} = \sigma(t^{\mathrm{S}} W_{\mathrm{ref}})$. The base-stream centroid serves as a detached soft target:
\begin{equation}
\mathcal{L}_{\mathrm{ref}}
=
(r_x^{\mathrm{S}} - c_{t,x}^{\mathrm{B}})^2
+
(r_y^{\mathrm{S}} - c_{t,y}^{\mathrm{B}})^2,
\end{equation}
encouraging the retrieved regions to follow the attention-induced visual focus of the original pretrained model.

The final objective combines the token-level cross-entropy loss from the SCVER stream with the alignment term:
\begin{equation}
\mathcal{L}
=
\mathcal{L}_{\mathrm{CE}}
+
\lambda \mathcal{L}_{\mathrm{ref}}.
\end{equation}

\section{Experiments}




\subsection{Implementation Details}
\label{sec:imple}
We evaluate VLMs with diverse encoder architectures and parameter scales (Table~\ref{tab:baseline}). During training, we freeze all pretrained backbone parameters and optimize only the components introduced by the state-aware attention module. The sampling module is applied every four layers, with further analysis in Sec.~\ref{sec:frequency}. Training data details are provided in Appendix~\ref{appendix:training data}. All efficiency measurements are conducted on NVIDIA H100 GPUs with a batch size of 1.

\begin{figure*}
  \centering
  \includegraphics[width=1.0\linewidth, trim=0 290 0 0, clip]{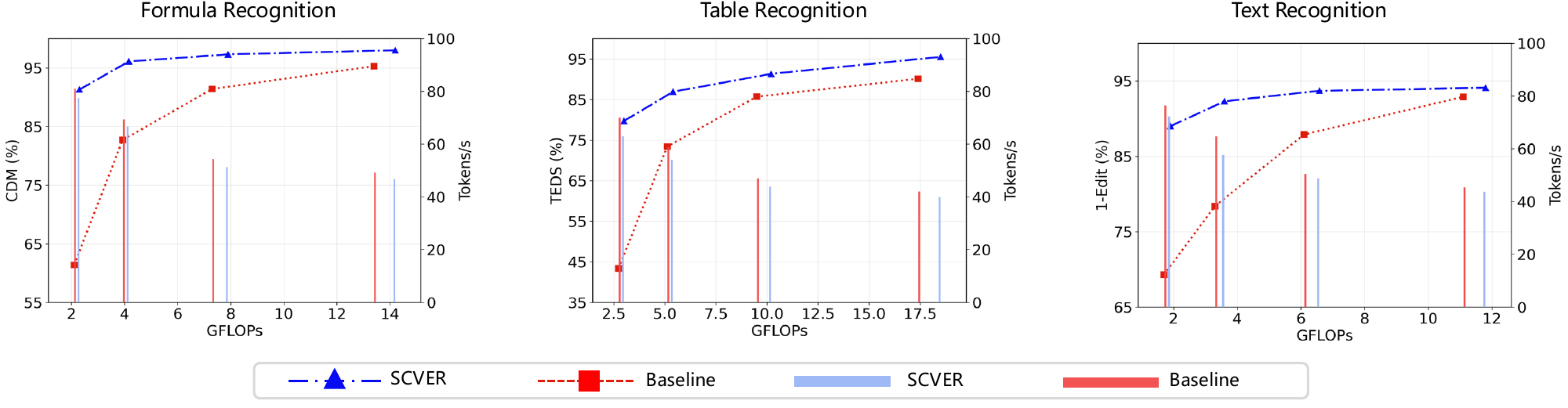}
  \caption{Accuracy and efficiency analysis. The x-axis shows input image size (0.4, 0.6, 0.8, 1.0) corresponding to per-token GFLOPs. Lines on the left y-axis indicate model performance and bars on the right y-axis indicate tokens per second. Integrating SCVER achieves a better accuracy-efficiency trade-off compared to the baseline (MinerU2.5).}
  \label{fig:11}
\end{figure*}

\subsection{Benchmarks and Evaluation Metrics}
We consider three representative categories:
(1) text recognition: B-MOD~\cite{kišš2019brnomobileocrdataset} and the OCR-block subset of DocLayNet;
(2) formula recognition (LaTeX): UniMER-Test~\cite{wang2024unimernetuniversalnetworkrealworld}, including Simple Printed Expressions (SPE), Complex Printed Expressions (CPE), Screen-Captured Expressions (SCE), and Handwritten Expressions (HWE);
(3) table structure recognition (HTML): PubTabNet~\cite{pubtab} and FinTabNet~\cite{fintab}.
To assess overall document parsing capability, we further report end-to-end page-level evaluation on OmniDocBench~\cite{ouyang2025omnidocbenchbenchmarkingdiversepdf}.
Since most document parsing systems do not rely on VLMs for layout detection, we defer layout detection experiments to Appendix~\ref{appendix:layout}.
We use Edit Distance (Edit) for text recognition, Character Detection Matching (CDM)~\cite{wang2024cdmreliablemetricfair} for formula recognition and Tree-Edit-Distance (TEDS)~\cite{pubtab} for table recognition.



\subsection{Main Results}
\label{sec:recognition}
Table~\ref{tab:recognition} compares SCVER with a broad range of methods on diverse document parsing benchmarks under both original and reduced input resolutions. We include specialized document recognition models such as DeepSeek-OCR-2~\cite{wei2026deepseekocr2visualcausal}, GLM-OCR~\cite{duan2026glmocrtechnicalreport}, and PaddleOCR-VL~\cite{cui2025paddleocrvlboostingmultilingualdocument}, as well as representative general-purpose VLMs~\cite{openai2024gpt4ocard,bai2025qwen3vltechnicalreport}.

Reducing input resolution is commonly used to control computational cost, but often leads to severe performance degradation. To evaluate robustness under such constraints, we integrate SCVER with the three backbone models in Sec.~\ref{sec:imple}. For Dolphin-1.5, results are reported at the original resolution, while for MinerU2.5-Pro and MonkeyOCR-Pro, we additionally evaluate reduced resolutions.

SCVER consistently improves performance across both standard and constrained settings. On Dolphin-1.5, it improves all benchmarks, e.g., PubTabNet from 85.9 to 91.9 and B-MOD from 82.7 to 94.0. More importantly, under reduced resolutions, SCVER significantly mitigates performance degradation. For example, at 0.5 resolution (about 25\% of the original visual tokens), MonkeyOCR-Pro drops from 87.4 to 58.3 on PubTabNet and from 83.3 to 62.4 on B-MOD, while SCVER recovers them to 92.8 and 91.3. A similar trend is observed on MinerU2.5-Pro, where SCVER improves PubTabNet from 82.0 to 95.0 and B-MOD from 73.3 to 94.7 at 0.7 resolution (about 49\% tokens).
These results demonstrate that SCVER reduces the reliance of VLMs on the initial input resolution.
 

\subsection{Accuracy--efficiency trade-off}
Figure~\ref{fig:11} analyzes the accuracy--efficiency trade-off under varying input resolutions. As input size decreases, the baseline suffers a rapid performance drop across all three tasks, indicating strong reliance on high-resolution inputs for preserving fine-grained visual details.

In contrast, SCVER consistently achieves higher accuracy at comparable or lower computation, with the largest gains appearing in low-compute regimes. At roughly $2$ GFLOPs, SCVER improves formula recognition from about $61\%$ to $91\%$, table recognition from about $43\%$ to $80\%$, and text recognition from about $69\%$ to $89\%$. The advantage remains substantial at moderate computation levels, where SCVER continues to outperform the baseline by around $10$--$13$ points across tasks.

Moreover, SCVER often reaches the baseline's high-cost performance at much lower computation. For example, in text recognition, SCVER achieves about $89\%$ at roughly $1.8$ GFLOPs, while the baseline requires more than $6$ GFLOPs to approach a similar level. Although SCVER introduces additional retrieval operations, the throughput remains comparable to the baseline, with only a small drop in tokens per second (as indicated by the bars), indicating that the practical efficiency is largely preserved.

\begin{table}[t]
\centering
\resizebox{\columnwidth}{!}{
\begin{tabular}{lcccc}
\toprule
 & \textbf{HWE} & \textbf{PubTabNet} & \textbf{B-MOD} & \textbf{FLOPs/Token} \\
\midrule
MinerU2.5-Pro & 95.3 & 90.1 & 85.6 & 1.15e+10 \\
\midrule
+VIVL & 94.1(\textcolor{red}{-1.2}) & 90.7(\textcolor{green}{+0.7}) &91.7(\textcolor{green}{+6.1}) & 1.19e+10 \\
+HVFA & 96.2(\textcolor{green}{+0.9}) & 91.4(\textcolor{green}{+1.3}) & 92.1(\textcolor{green}{+6.5}) & 1.34e+10 \\
+SCVER (Ours) & 97.4(\textcolor{green}{+2.1}) & 95.2(\textcolor{green}{+5.1}) & 95.1(\textcolor{green}{+9.5}) & 1.21e+10 \\
\bottomrule
\end{tabular}
}
\caption{Comparison of different methods. 
}
\label{tab:compare}
\end{table}

\subsection{Comparison with Existing Methods}
\label{sec:compare}
We compare our method with VIVL~\cite{Cao_2024_CVPR} and HVFA~\cite{park2024hierarchicalvisualfeatureaggregation}, both of which enhance visual representations before decoding, whereas our method dynamically retrieves fine-grained visual evidence during decoding.
As shown in Table~\ref{tab:compare}, we conduct experiments using MinerU2.5 as the backbone model. VIVL introduces the smallest computational overhead but provides limited performance gains and even slightly degrades text recognition accuracy, possibly because it mainly emphasizes semantic-level feature fusion rather than improving access to fine-grained visual details. HVFA improves performance across datasets but incurs the largest computational cost. In comparison, our method achieves the best overall performance with only a small increase in computation, demonstrating the effectiveness of dynamic visual evidence retrieval for enhancing fine-grained perception.

\subsection{Generalization to Text-Rich VQA}
Text-rich VQA also requires models to identify fine-grained, answer-relevant visual evidence from complex images. Since attention cost grows quadratically with the number of visual tokens, reducing image resolution is a common way to improve efficiency, but it may also remove local details needed for answering questions.

As shown in Table~\ref{tab:docvqa_resolution}, we further evaluate SCVER on DocVQA\cite{mathew2021docvqadatasetvqadocument}, TextVQA\cite{singh2019vqamodelsread}, and InfoVQA\cite{mathew2021infographicvqa} using Qwen2.5-VL-7B as the backbone. We compare with DocVLM~\cite{nacson2024docvlmmakevlmefficient}, which introduces an additional OCR module to compensate for information loss under low-resolution inputs. Under a limited token budget of 320, SCVER remains competitive with DocVLM on DocVQA and TextVQA, and performs better on InfoVQA. Under 576 tokens, both methods improve over the backbone, with comparable overall performance.

Although SCVER does not consistently outperform the OCR-augmented baseline, it achieves competitive results without relying on an external OCR system. In contrast, DocVLM is specifically designed for OCR-assisted document question answering, while SCVER operates as a model-internal, decoding-time retrieval mechanism that can be directly applied across different vision-language settings. This suggests that state-conditioned visual evidence retrieval is not limited to document parsing, and can also benefit broader text-rich grounded understanding tasks.

\begin{table}[t]
\centering
\resizebox{\columnwidth}{!}{
\begin{tabular}{lcccc}
\toprule
Method & Tok & DocVQA & TextVQA & InfoVQA \\
\midrule
Qwen2.5-VL      & 576 & 92.7 & 80.9 & 71.6 \\
\midrule
DocVLM     & 320 & 92.3(\textcolor{red}{-0.4}) & 79.7(\textcolor{red}{-1.2}) & 67.3(\textcolor{red}{-4.3}) \\
SCVER (Ours)& 320 & 90.4(\textcolor{red}{-2.3}) & 78.0(\textcolor{red}{-2.9}) & 69.4(\textcolor{red}{-2.2}) \\
\midrule
DocVLM     & 576 & 94.2(\textcolor{green}{+1.5}) & 83.8(\textcolor{green}{+2.9}) & 72.8(\textcolor{green}{+1.2}) \\
SCVER (Ours)& 576 & 93.5(\textcolor{green}{+0.8}) & 82.9(\textcolor{green}{+2.0}) & 73.1(\textcolor{green}{+1.5}) \\
\bottomrule
\end{tabular}
}
\caption{
Results on document VQA benchmarks under different visual token budgets.
\textbf{Tok} denotes the maximum number of visual tokens after image resizing.
}
\label{tab:docvqa_resolution}
\end{table}

\begin{figure}[t]
  \centering
  \includegraphics[width=1.0\linewidth, trim= 0 0 0 0, clip]{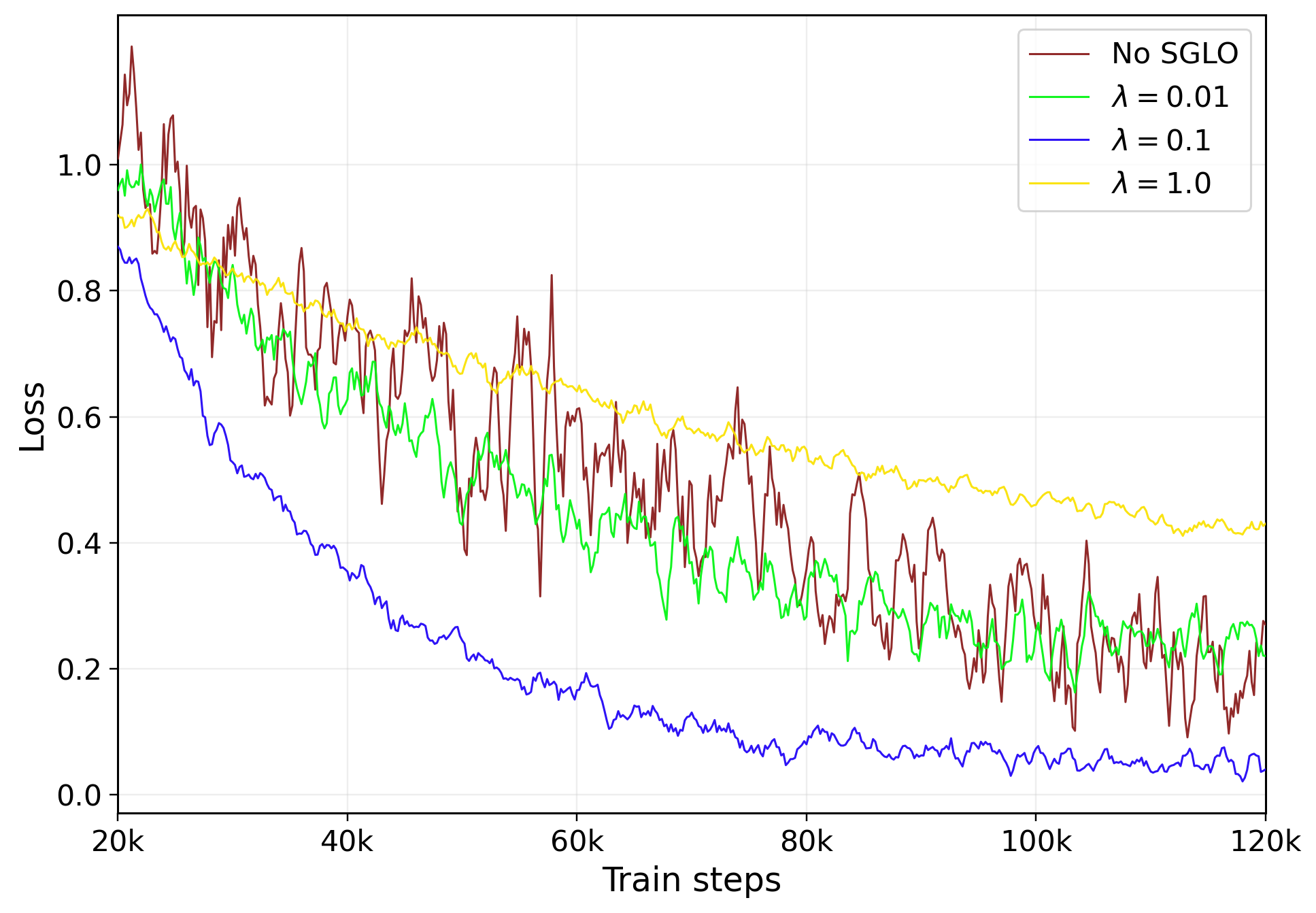}
  \caption{Training loss curves with and without the Spatially-Guided Learning Objective (SGLO). $\lambda$ denotes the weight of the SGLO term.}
\label{fig:6}
\end{figure}

\subsection{Ablation Study}

In the main paper, we analyze the scaling behavior of state-conditioned visual evidence
retrieval and the effect of the Spatially-Guided Learning Objective. Further analyses of other design choices and whether the observed performance gains arise from data scaling effects are provided in Appendix~\ref{appendix:supp}.

\begin{figure*}[ht]
  \centering
  \includegraphics[width=1.0\linewidth, trim=0 405 315 0, clip]{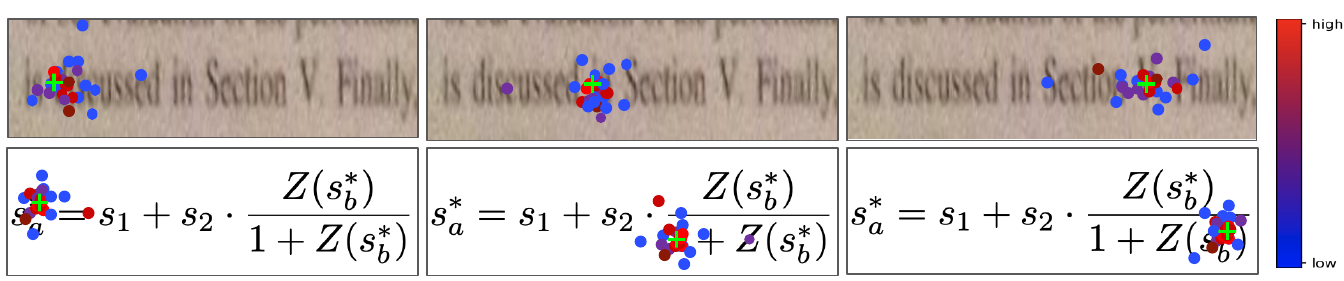}
  \caption{
\textbf{Visualization of state-conditioned retrieval points.}
Multi-head retrieval points (after applying offsets to the reference point) and their attention weights. Each circle represents a retrieval point, with color indicating its attention weight.}
  \label{fig:ablation2}
\end{figure*}

\subsubsection{Effect of Training Strategy}
\label{sec:train}

Directly applying SCVER to VLMs yields only limited performance
improvements (Table~5, first two rows), as learning state-conditioned
visual retrieval is challenging and often leads to unstable optimization.
Under token-level supervision alone, the model must learn where to retrieve
visual evidence from indirect and locally sampled gradients. To address
this issue, SGLO uses visual-token attention centroids produced by an independent no-gradient frozen-base
stream as online, token-conditioned spatial targets
to guide the retrieval process.

With SGLO, the advantage of SCVER becomes significantly more pronounced. For example, on PubTabNet, the performance gain increases from 2.5\% to 6.9\%. Training dynamics further show that SGLO stabilizes optimization (Fig.~\ref{fig:6}). Without SGLO, training exhibits noticeable oscillations. A small weight ($\lambda=0.01$) provides limited stabilization, while a large weight ($\lambda=1.0$) overly constrains optimization and slows convergence. Empirically, $\lambda=0.1$ achieves a better balance between stability and performance.

Overall, SGLO effectively regularizes the retrieval process, improving both training stability and downstream performance.

\begin{table}[t]
\centering
\resizebox{\columnwidth}{!}{
\begin{tabular}{ccccc}
\toprule
\textbf{\textit{SCVER}} & \textbf{\textit{SGLO}} & \textbf{SCE} & \textbf{PubTabNet} & \textbf{B-Mod} \\
\midrule
 &  & 95.1 & 85.9 & 82.7 \\
\checkmark &  & 95.4 & 88.1 & 85.9 \\
\checkmark & \checkmark & \textbf{97.6} & \textbf{91.9} & \textbf{94.0} \\
\bottomrule
\end{tabular}
}
\caption{Ablation study on the effect of SGLO. 
}
\label{tab:sglo}
\end{table}

\subsubsection{Effect of SCVER Application Frequency}
\label{sec:frequency}

We further analyze the effect of the SCVER module frequency on both accuracy and computational cost. Experiments are conducted using MinerU2.5~\cite{niu2025mineru25decoupledvisionlanguagemodel} as the backbone model on the PubTabNet~\cite{pubtab}.

\begin{figure}[t]
  \centering
  \includegraphics[width=1.0\linewidth, trim= 0 130 330 0, clip]{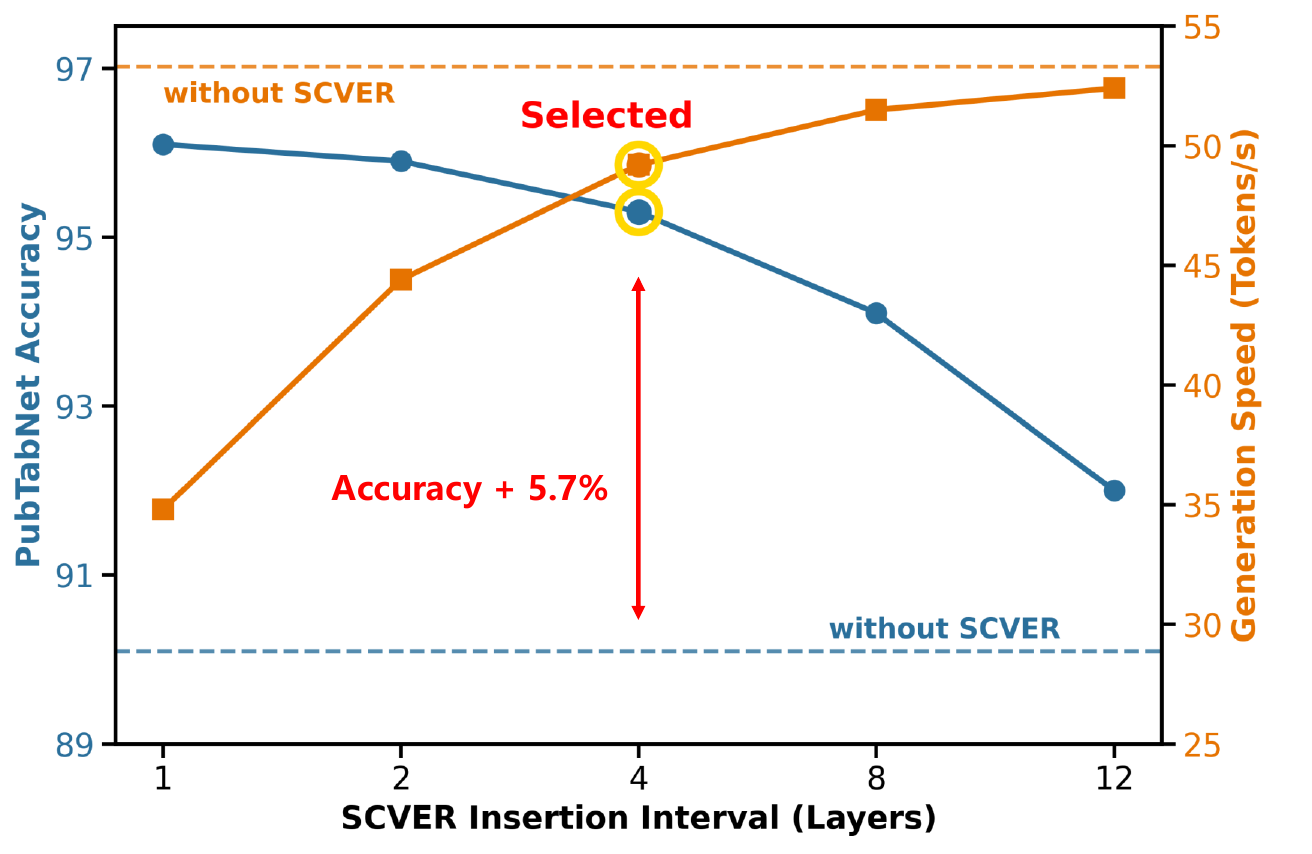}
  \caption{Effect of retrieval module application frequency on accuracy. The x-axis indicates the interval of decoder layers at which sampling module is inserted.}
\label{fig:5}
\end{figure}

As shown in Fig.~\ref{fig:5}, increasing the interval between SCVER applications significantly improves decoding speed, while the accuracy decreases gradually. This indicates that applying SCVER too frequently yields diminishing returns in performance.

Based on this trade-off between efficiency and performance, we apply SCVER every four layers. Compared with applying it at every decoder layer, this configuration results in only a 0.7\% drop in accuracy while reducing computational cost by approximately 10\%. This setting achieves a better balance between efficiency and performance.

\subsection{Case study.}

We analyze the behavior of SCVER using a representative example in Fig.~\ref{fig:ablation2}, which visualizes retrieved locations and their weights for different tokens in text and formula regions. The retrieved points are highly localized and vary across tokens: tokens requiring fine-grained visual details, such as mathematical symbols or specific characters, focus on corresponding local regions, while different tokens attend to distinct spatial areas. This demonstrates that SCVER performs token-specific, state-dependent visual evidence retrieval, enabling dynamic access to relevant fine-grained information during decoding.

\section{Conclusion}

In this paper, we introduce State-Conditioned Visual Evidence Retrieval (SCVER), a new paradigm for fine-grained visual perception in vision--language models (VLMs). Instead of relying on static visual representations that must encode both coarse structure and fine-grained details, SCVER formulates perception as the retrieval of token-specific visual evidence conditioned on the current decoding state. To realize this paradigm, we combine a coarse-to-fine retrieval design with a Spatially-Guided Learning Objective (SGLO), enabling efficient access to fine-grained visual evidence during autoregressive decoding. Extensive experiments show that SCVER consistently improves robustness and achieves a better accuracy--efficiency trade-off on document parsing benchmarks, especially under reduced input resolution. Overall, these results suggest that decoding-time visual evidence retrieval provides a practical and effective direction for enhancing fine-grained perception in future VLMs.

\section*{Limitations}
The proposed SGLO training strategy requires an additional no-gradient frozen-base forward and explicit visual-token attention maps from selected decoder layers. During this auxiliary forward, these layers must expose their attention weights through an eager attention path, rather than relying entirely on optimized implementations such as FlashAttention or SDPA. Although the frozen-base and SCVER streams share the same backbone weights and do not require a separate teacher model, the additional forward pass and attention extraction increase training-time computation and may reduce training throughput. This auxiliary stream is disabled after training: inference uses only the SCVER-augmented stream and therefore incurs no SGLO-specific inference overhead.

\section*{Acknowledgments}
The authors wish to thank the anonymous reviewers for their helpful comments and WisPaper for its assistance in literature retrieval. This work was partially funded by Henan Province Major Industrial Challenge-Based Innovation (No.251000210300), National Natural Science Foundation of China (No.62476061, 62376061, 62576106).

\bibliography{custom}

\clearpage
\appendix

\section{Training Data}
\label{appendix:training data}

Our training data are mainly derived from the training splits corresponding to the evaluation benchmarks described in the main text. These datasets provide diverse supervision signals for layout understanding, formula recognition, and text recognition.

However, most existing open-source document parsing datasets are primarily constructed from English documents. In contrast, the OmniDoc benchmark contains a large number of non-English document elements, such as Chinese formulas and tables, while it does not provide an official training set. This mismatch between training resources and evaluation data may limit the robustness of models in multilingual document scenarios.

To address this issue, we supplement additional multilingual training resources and further refine existing datasets. Specifically, we construct additional data for formula recognition and text recognition to better support multilingual document parsing tasks.

\subsection{Formula Recognition Dataset}

Existing open-source formula datasets are typically constructed by extracting mathematical expressions from a large number of arXiv files and re-rendering them into image–LaTeX pairs. Although this pipeline largely avoids syntactic errors, variations in authors’ writing styles often lead to multiple LaTeX representations of the same mathematical expression. Such diversity introduces semantic ambiguities and inconsistencies in the ground truth.

To mitigate this issue, we identify common divergent LaTeX representations and standardize them to ensure semantic and structural consistency. After normalization, the standardized LaTeX code is re-rendered to verify correctness, producing a cleaner and more coherent dataset whose input images and output LaTeX sequences are strictly aligned.

In addition, to better support multilingual document parsing scenarios such as OmniDoc, we incorporate an additional 0.5M multilingual formula samples from the open-source project oleehyo\_latex\_formulas\_80M, which provides formula expressions in multiple languages.

\begin{table}[ht]
  \centering
  \label{stand}
  \begin{tabular}{lcc}
    \toprule
    \textbf{Issue} & \textbf{Original} & \textbf{Standardized} \\
    \hline
    Bracket        & \texttt{\textbackslash\{}                 & \texttt{\textbackslash lbrace} \\
    Subsup         & \texttt{a\textasciicircum 1\_2}           & \texttt{a\_2\textasciicircum 1} \\
    Prime          & a$'$                                      & \texttt{a\textasciicircum{\{\textbackslash prime\}}} \\
    Fraction       & \texttt{\textbackslash over}              & \texttt{\textbackslash frac} \\
    Space          & \texttt{\textbackslash array\{l c\}}      & \texttt{\textbackslash array\{lc\}} \\
    \toprule
  \end{tabular}
  \caption{Examples of LaTeX standardization for various symbols and expressions.}
\end{table}

\subsection{Text Recognition Dataset}

Our text recognition data mainly come from two sources. The first is DocLayNet~\cite{doclaynet2022}, which provides fine-grained layout annotations with 11 classes across six document categories. Among these labels, mathematical content is represented by a general Formula category, without a separate label for inline formulas. This granularity is insufficient for our scientific-document text recognition setting, where inline mathematical expressions are often embedded within textual regions.

To complement this limitation, we construct a scientific-document-oriented paragraph recognition dataset based on the Text\_Completion\_arXiv corpus, a page-level annotated dataset of arXiv papers. Building upon these annotations, we generate large-scale block-level image–text pairs using a combination of detection and recognition models supported by heuristic rules.

Specifically, we first employ a layout detection model to segment page-level images into individual layout elements and crop the corresponding regions using detected bounding boxes. Each cropped region is then processed by a state-of-the-art text recognition model to obtain textual predictions. Meanwhile, paragraph-level ground truth is extracted from the original page-level text using \texttt{\textbackslash n\textbackslash n} as separators. Finally, we align the recognized text with extracted paragraphs using edit-distance filtering and order-aware matching with strict thresholds to ensure correctness.

Through this pipeline, we construct a large-scale, high-quality element-level dataset tailored for robust scientific document text recognition.
\section{Supplementary experiments}
\label{appendix:supp}

\begin{table*}[t]
\centering
\resizebox{\textwidth}{!}{
\begin{tabular}{lcccccc}
\toprule
Method & Textual & Image & Table & Equation & Page Margins & Full Page \\
\midrule
\multicolumn{6}{c}{D4LA} \\
\midrule
DocLayout-YOLO~\cite{zhao2024doclayoutyoloenhancingdocumentlayout} & 90.8 & 62.6 & 89.8 & 91.1 & - &88.5 \\
PP-StructureV3~\cite{cui2025paddleocr30technicalreport} & 90.0 & 67.9 & 89.7 & 92.1 & 79.1 &88.3 \\
MinerU2.5~\cite{niu2025mineru25decoupledvisionlanguagemodel} & 94.6 & 72.8 & 91.4 & 91.0 & 84.2 &91.4\\
\rowcolor{gray!30}
MinerU2.5+SCVER & 95.8 & 74.9 & 93.2 & 92.9 & 88.5 &93.2 \\
\midrule
\multicolumn{6}{c}{DocLaynet} \\
\midrule
DocLayout-YOLO~\cite{zhao2024doclayoutyoloenhancingdocumentlayout} & 91.2 & 91.3 & 94.8 & 82.8 & - &92.0\\
PP-StructureV3~\cite{cui2025paddleocr30technicalreport} & 93.8 & 94.2 & 96.7 & 92.1 & 77.4 &94.0\\
MinerU2.5~\cite{niu2025mineru25decoupledvisionlanguagemodel} & 94.8 & 95.9 & 97.1 & 93.5 & 86.3 &95.2\\
\rowcolor{gray!30}
MinerU2.5+SCVER & 95.2 & 96.7 & 97.6 & 94.7 & 89.7 &95.5\\
\bottomrule
\end{tabular}
}
\caption{Comparison of layout analysis performance across different methods and content types on multiple benchmark datasets.}
\label{tab:detection_appendix}
\end{table*}

\begin{figure*}
  \centering
  \includegraphics[width=1.0\linewidth, trim=0 0 0 0, clip]{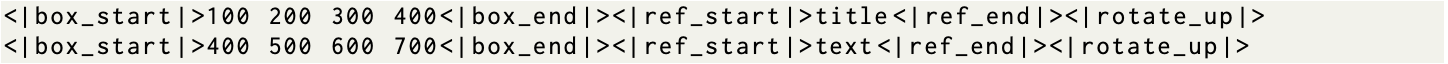}
  \caption{Output format of layout detection.}
  \label{fig:detection_output_format}
\end{figure*}

\subsection{Training Details}
We conducted all training and inference speed evaluations using the PyTorch framework on NVIDIA H100 GPUs. During training, we used an initial learning rate of $1\mathrm{e}{-5}$ and employed a cosine learning rate scheduler to progressively adjust the model parameters.

\subsection{Page-level Evaluation Metrics}

We conduct page-level evaluation on the OmniDocBench benchmark. To eliminate the influence of layout detection errors on downstream evaluation, the page-level assessment focuses on three major document layout elements: text, tables, and formulas.

The overall page-level score is computed by averaging the performance on these three elements. Specifically, the evaluation uses Edit Distance for text, TEDS for tables, and CDM for formulas. The overall score is calculated as follows:

\begin{equation*}
\resizebox{\columnwidth}{!}{$
\text{Overall} =
\frac{(1 - \text{Text}^{\text{Edit}}) \times 100 + \text{Table}^{\text{TEDS}} + \text{Formula}^{\text{CDM}}}{3}
$}
\end{equation*}


\subsection{Layout Detection}
\label{appendix:layout}

In the main paper, we observe that incorporating the proposed State-Conditioned Visual Evidence Retrieval (SCVER) paradigm into VLMs leads to only marginal improvements on page-level document parsing benchmarks. A key reason is that many document parsing systems rely on additional layout detection modules, where the final parsing results are largely determined by the outputs of these detectors. Since layout detection itself requires fine-grained visual perception, improvements brought by SCVER may not be fully reflected in page-level evaluations.

To further evaluate the effectiveness of SCVER in layout detection tasks, we adapt SCVER to the MinerU2.5 model and conduct experiments on the publicly available DocLayNet~\cite{doclaynet2022} and D$^4$LA~\cite{da2023visiongridtransformerdocument} datasets. We compare our approach with representative layout detection methods, including the YOLO-based DocLayout-YOLO~\cite{zhao2024doclayoutyoloenhancingdocumentlayout} and the Detection Transformer-based PP-StructureV3. Since the two datasets adopt different label taxonomies, we follow the label mapping strategy proposed in the MinerU2.5 report~\cite{niu2025mineru25decoupledvisionlanguagemodel} to convert the annotations into a unified label space.

Table~\ref{tab:detection_appendix} presents detailed results across different layout element categories. The results show that the vision–language model MinerU2.5 significantly outperforms traditional detection models such as YOLO, as well as attention-based detection architectures like Detection Transformer. We attribute this advantage to the joint training of MinerU2.5 on both layout detection and layout element recognition tasks. This joint learning allows the model to implicitly leverage textual modality information during layout detection, enabling not only more accurate bounding box localization but also substantially improved category prediction.

Building upon this strong baseline, the proposed SCVER further enhances the model's ability to utilize fine-grained visual information during decoding. As a result, the generative vision–language model can better capture detailed visual structures, leading to additional improvements in layout detection performance.

\begin{table*}[t]
\centering
\resizebox{\textwidth}{!}{%
\begin{tabular}{l c c c c c c c c c c}
\toprule

\multirow{2}{*}{\textbf{Method}} &
\multicolumn{4}{c}{\textbf{Formula$^{\mathbf{CDM} \uparrow}$}} &
\multicolumn{2}{c}{\textbf{Table$^{\mathbf{TEDS} \uparrow}$}} &
\multicolumn{2}{c}{\textbf{Text$^{\mathbf{Edit} \downarrow}$}} \\
\cmidrule(lr){2-5}  \cmidrule(lr){6-7} \cmidrule(lr){8-9}
& SPE& SCE & CPE & HWE & PubTabNet & FinTabNet & DocLaynet & B-MOD \\
\midrule
Dolphin-1.5   & 97.8 & 95.1 & 87.7& - & 85.9 & 87.4 & 0.017 & 0.304         \\
Dolphin-1.5*         & 97.9 & 95.3 & 87.4& - & 86.1 & 87.0 & 0.014 & 0.244         \\
Dolphin-1.5 + SCVER   & 98.4 & 97.6 & 90.2& - & 91.9 & 92.4 & 0.011 & 0.068          \\

\midrule
MonkeyOCR-Pro  & 97.6 & 94.9 & 91.4 & 92.2 & 87.4 & 86.4 & 0.020 & 0.204 \\
MonkeyOCR-Pro*      & 97.4 & 95.3 & 91.7 & 92.0 & 88.0 & 86.2 & 0.018 & 0.154 \\
MonkeyOCR-Pro + SCVER  & \textbf{99.2} & 97.2 & 93.9 & 97.2 & \textbf{95.2} & 91.4 & 0.014 & \textbf{0.028}  \\

\midrule
MinerU2.5     & 98.4 & 96.4 & 96.6& 94.4 & 89.1 & 95.6 & 0.013 &0.218  \\
MinerU2.5*             & 98.6 & 96.1 & 96.9& 93.9 & 89.2 & 95.8 & 0.011 &0.158 \\
MinerU2.5 + SCVER      & \textbf{99.2} & \textbf{98.7} & \textbf{97.1}& \textbf{97.4} & 95.1 & \textbf{95.9} & \textbf{0.010} &0.052 \\
\bottomrule

\end{tabular}%
} 
\caption{Analysis of potential data leakage to verify whether the observed performance gains are caused by training data overlap. The symbol * indicates that the model is further fine-tuned on the same data used to train SCVER.}
\label{tab:data_leak}
\end{table*}

\begin{table}[t]
\centering
\resizebox{\columnwidth}{!}{
\begin{tabular}{ccccccc}
\toprule
\multicolumn{3}{c}{Configuration} & \multicolumn{3}{c}{Performance} & Efficiency \\
\cmidrule(lr){1-3} \cmidrule(lr){4-6} \cmidrule(lr){7-7}
H & P & F & Formula & Table & Text & FLOPs \\
\midrule
8  & 2 & 1 & 95.6 & 87.6 & 0.028 & 9.58e+08 \\
16 & 4 & 1 & 97.3 & 91.9 & 0.011 & 9.90e+08 \\
24 & 8 & 2 & 97.6 & 92.4 & 0.011 & 1.27e+09 \\
\bottomrule
\end{tabular}
}
\caption{Performance and efficiency under different configurations. \( H \), \( P \), and \( F \) correspond to the number of heads, sampling points, and high-scale feature maps used in the SCVER computation.}
\label{tab:efficiency_append}
\end{table}

\subsection{Internal Hyperparameters of SCVER}
\label{appendix:hyper}
In the main paper, we primarily investigate the impact of the application frequency of SCVER on model accuracy and inference efficiency. Here, we further explore several internal hyperparameters of SCVER, including the number of high-resolution feature maps used, the number of attention heads, and the number of sampling points per head.

In our previous experiments, MinerU2.5 is built upon the Qwen2VL architecture~\cite{wang2024qwen2vlenhancingvisionlanguagemodels}, which employs a NaViT-based visual encoder. This encoder only provides a single high-resolution feature map before downsampling, limiting our ability to study the influence of multi-scale visual features. To conduct a more comprehensive analysis, we instead adopt Dolphin1.5, which utilizes a hierarchical visual encoder and thus allows us to better explore these hyperparameters.

As shown in Table~\ref{tab:efficiency_append}, $F$ denotes the number of high-resolution feature maps used by SCVER. When $F=1$, the model uses a $2\times2$ high-resolution feature map, while $F=2$ indicates the use of both $2\times2$ and $4\times4$ feature maps. The results show that when $F=1$, increasing the number of sampling points per head is already sufficient to approach the peak performance. Incorporating additional high-resolution feature maps brings only marginal accuracy gains while significantly increasing computational cost. This observation suggests that the key benefit of SCVER mainly comes from dynamic, state-conditioned sampling rather than simply aggregating more high-resolution visual features.

\begin{table}[t]
\centering
\begin{tabular}{lccc}
\toprule
$\lambda$ & Formula & Table & Text \\
\midrule
0    & 96.4 & 95.6 & 0.218 \\
0.01 & 96.7 & 95.3 & 0.166 \\
0.1 (selected) & \textbf{98.7} & \textbf{95.9} & 0.028 \\
1.0  & 98.2 & 95.7 & \textbf{0.017} \\
\bottomrule
\end{tabular}
\caption{Performance under different $\lambda$ values on SGLO.}
\label{tab:lambda_ablation}
\end{table}

\subsection{Impact of Weight Coefficients on SGLO}
\label{appendix:Coefficients}
As described earlier, we introduce the State-Conditioned Visual Evidence Retrieval (SCVER) module to enable state-conditioned dynamic fine-grained perception in vision–language models. However, when integrating SCVER into VLMs, the conventional token-level loss cannot provide direct supervision for the sampling locations, which limits the performance gains. Visualization results further reveal that the predicted sampling points tend to collapse into nearly uniform global sampling patterns.

To address this issue, we introduce the Spatially-Guided Learning Objective (SGLO) as an additional loss term to provide explicit guidance for the sampling process. We further conduct an ablation study on the weight of this loss term, as shown in Table~\ref{tab:lambda_ablation}. When the weight of SGLO is too small, the performance gain remains marginal, suggesting that the guidance signal is insufficient to effectively guide the sampling behavior. In contrast, assigning an excessively large weight to this loss term may dominate the training objective, potentially constraining the model's ability to learn semantic representations. Consequently, a balanced weighting between the two objectives is necessary to achieve the best performance.

\begin{figure*}
  \centering
  \includegraphics[width=0.8\linewidth, trim=0 0 0 0, clip]{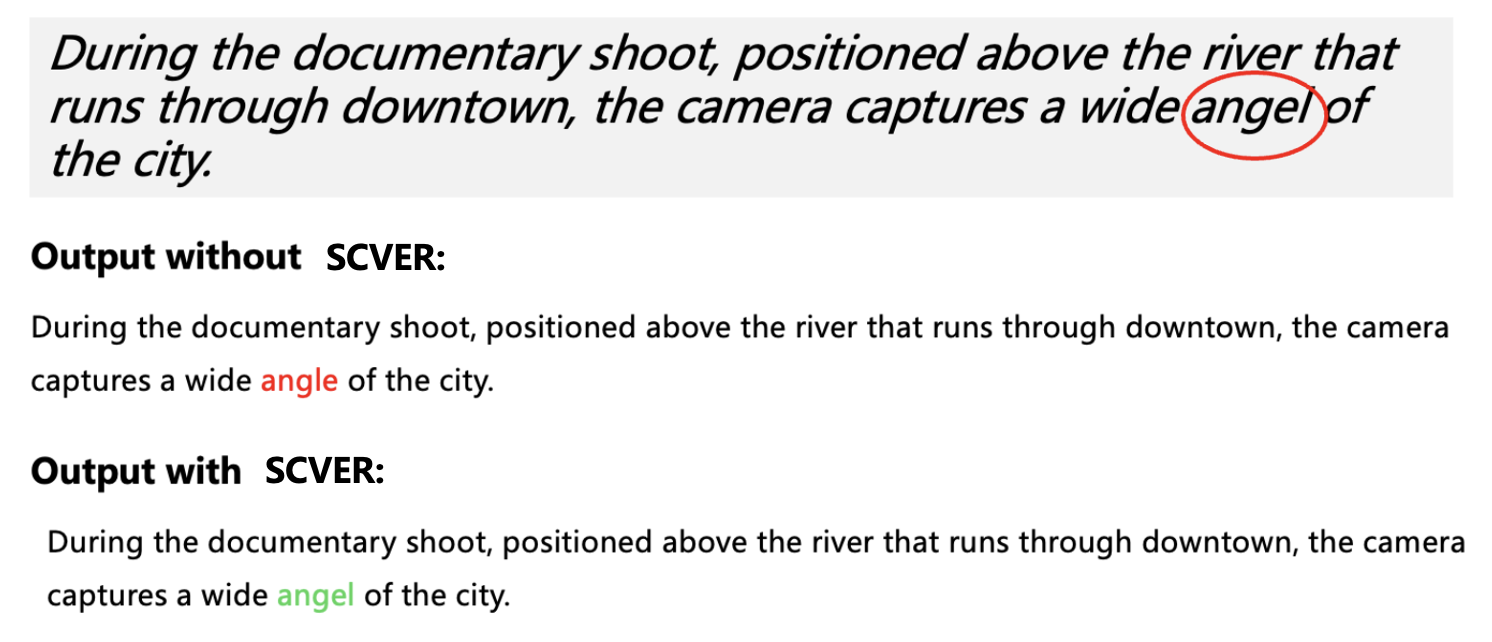}
  \caption{Comparison between the base model and the model equipped with the SCVER module on text recognition. The base model tends to rely on the decoder's language prior, while the model with SCVER follows the visual evidence more faithfully.}
  \label{fig:huanjue}
\end{figure*}

\subsection{Data Leakage Ablation Study}
\label{appendix:leakage}
Since the training data of the backbone model are not publicly available, it is difficult to determine whether the performance improvements observed in our experiments originate from the proposed design or from potential distribution differences between the backbone model’s training data and the evaluation datasets. To investigate this factor, we conduct additional experiments to analyze the impact of such data-related effects.

As shown in Table~\ref{tab:data_leak}, fine-tuning the backbone model with these datasets results in only limited improvements. Except for a moderate gain on the B-MOD dataset, most benchmarks show negligible changes, and some even exhibit slight performance degradation. This suggests that simply increasing the amount of training data does not consistently lead to performance improvements.

Furthermore, even on B-MOD, where visual quality is relatively degraded, the performance gains obtained from additional training data remain smaller than those achieved by introducing our perception mechanism. These observations indicate that the improvements observed in our experiments are unlikely to be explained solely by data scaling. Instead, they highlight the importance of explicitly modeling fine-grained visual perception in generative vision–language models.

\section{Study Case}
\label{appendix:study case}

In the last few pages of the appendix, we present additional examples demonstrating the accuracy gains brought by the SCVER module through enhanced fine-grained visual perception. During the analysis of failure cases, we also observed an interesting counterexample. As shown in Fig.~\ref{fig:huanjue}, we initially assumed that the base model made an error in text recognition. However, upon closer inspection, we found that the spelling in the original image was actually incorrect, while the base model produced a semantically plausible output. According to previous studies, this phenomenon may be attributed to the strong language priors accumulated by the decoder during large-scale pretraining. When encountering ambiguous or uncertain visual inputs, the model's predictions may be dominated by the decoder's language modeling capability, resulting in outputs that conform to linguistic statistics rather than strictly following the visual evidence.

In contrast, when the model is equipped with the SCVER module, this behavior appears to change. The model tends to produce outputs that more faithfully follow the visual information in the image rather than relying on language priors for semantic correction. This observation suggests that the SCVER module not only improves the fine-grained visual perception ability of vision–language models but may also potentially mitigate hallucination caused by language-prior dominance, thereby encouraging the model to better adhere to visual evidence.

Nevertheless, it should be noted that this issue is not the primary focus of our work, and due to space limitations we do not provide a systematic investigation of this phenomenon. Moreover, from the perspective of text recognition tasks, it remains an open question whether models should possess the ability to automatically correct spelling errors or instead strictly reproduce the text as it appears in the image. We therefore leave this interesting observation for future exploration.

\begin{figure*}
  \centering
  \includegraphics[width=0.9\linewidth, trim=0 0 430 0, clip]{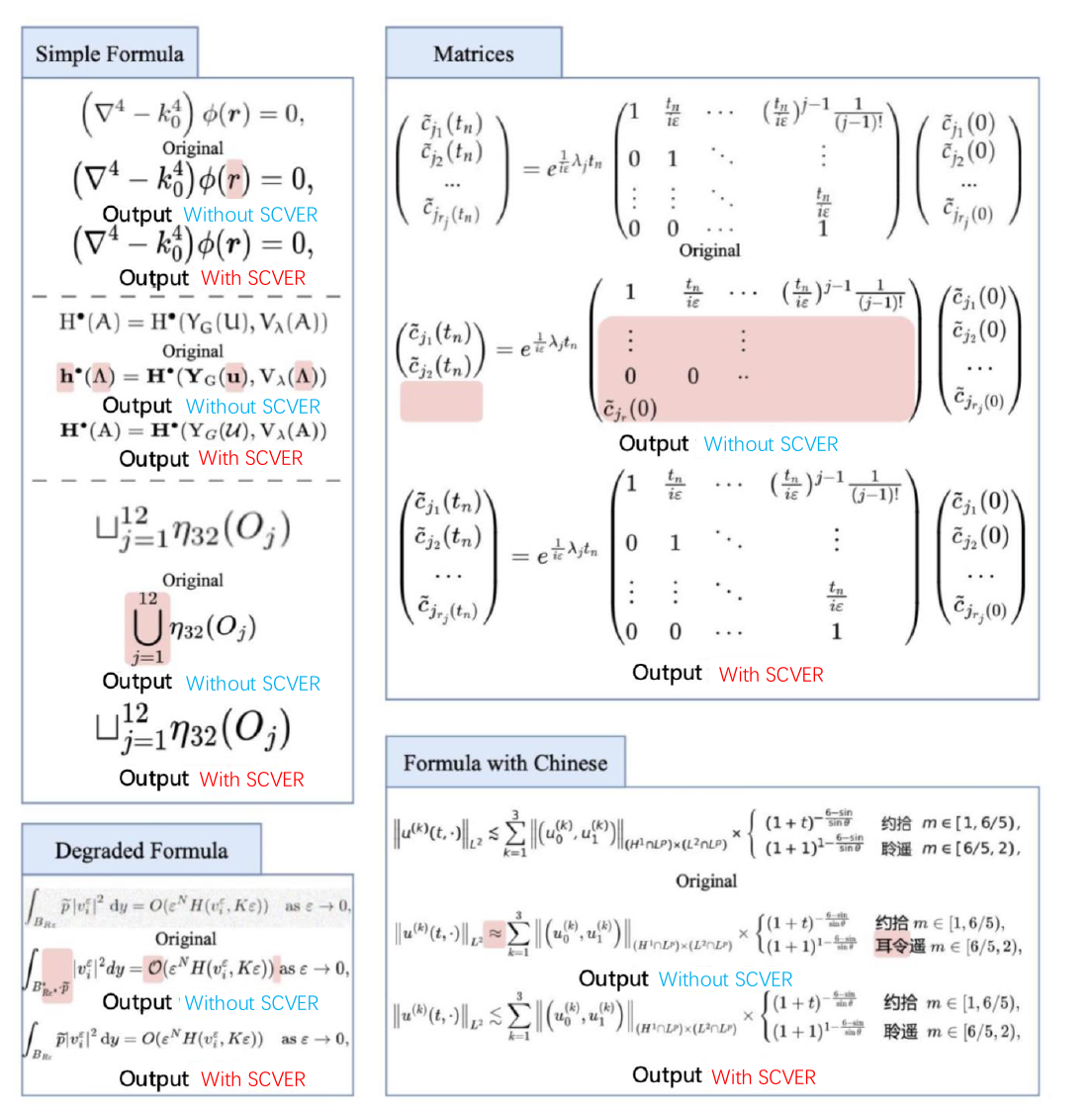}
  \caption{Impact of the SCVER strategy on formula recognition. This figure compares model outputs with and without the SCVER module
across various types of mathematical expressions, including simple formulas, matrices, degraded formulas, and formulas containing Chi-
nese text. The highlighted regions indicate the positions where the model without SCVER fails to preserve structural correctness or symbol
fidelity. These comparisons demonstrate that SCVER effectively mitigates such errors by maintaining fine-grained structural details and
improving symbol-level accuracy, resulting in outputs that more closely align with the original expressions.}
\end{figure*}

\begin{figure*}
  \centering
  \includegraphics[width=\linewidth, trim=0 210 0 0, clip]{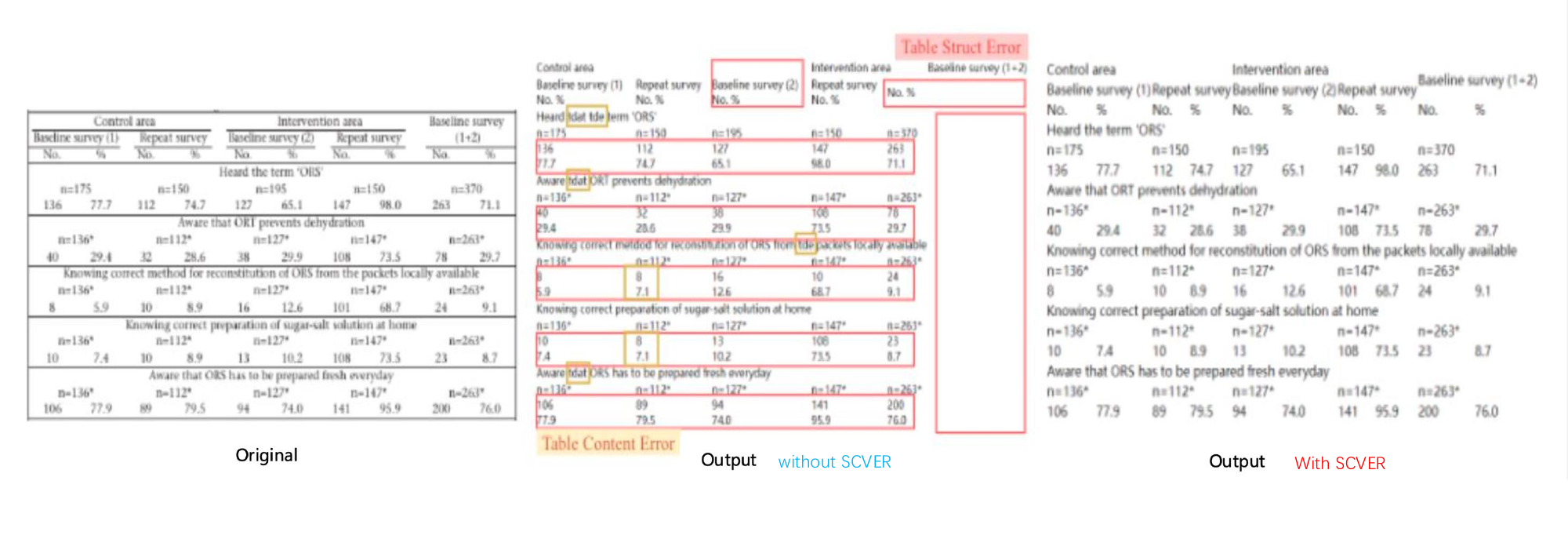}
  \caption{Impact of the SCVER strategy on table recognition.This figure compares model outputs with and without the SCVER module on
a complex table containing multi-row, multi-column, and merged-cell structures. The highlighted regions indicate structural and content
errors produced by the model without SCVER, including misaligned cell boundaries, incorrect row–column associations, and incorrectly
recognized text. These comparisons demonstrate that SCVER effectively preserves table-structure integrity and improves cell-level content
fidelity, enabling the model to generate outputs that more accurately reflect the original tabular layout and data.}
\end{figure*}
\end{document}